\documentclass[runningheads]{llncs}

\usepackage[titletoc]{appendix}
\usepackage{graphicx}
\usepackage{amsmath,amssymb} 
\usepackage{rotating}
\usepackage{multirow}
\usepackage{bm}
\usepackage[ruled]{algorithm2e}

\usepackage{times}
\usepackage{epsfig}
\usepackage{mathtools}
\usepackage{enumitem}
\usepackage[usenames,dvipsnames]{xcolor}

\begin{document}

\def\bigO2{\mbox{${\cal O}$}}
\def\bigO{O}

\def\mA{\mathcal{A}}
\def\mB{\mathcal{B}}
\def\mC{\mathcal{C}}
\def\mG{\mathcal{G}}
\def\mV{\mathcal{V}}
\def\mE{\mathcal{E}}
\def\mF{\mathcal{F}}
\def\mH{\mathcal{H}}
\def\mL{\mathcal{L}}
\def\mN{\mathcal{N}}
\def\mS{\mathcal{S}}
\def\mT{\mathcal{T}}
\def\mW{\mathcal{W}}
\def\mX{\mathcal{X}}
\def\mY{\mathcal{Y}}
\def\1n{\mathbf{1}_n}
\def\0{\mathbf{0}}
\def\1{\mathbf{1}}

\def\balpha{\boldsymbol{\alpha}}
\def\bdelta{\boldsymbol{\delta}}
\def\bzeta{\boldsymbol{\zeta}}
\def\bphi{\boldsymbol{\phi}}
\def\btau{\boldsymbol{\tau}}
\def\bmu{\boldsymbol{\mu}}
\def\bsigma{\boldsymbol{\sigma}}
\def\bSigma{{\bm \Sigma} }
\def\btheta{\boldsymbol{\theta}}
\def\bvartheta{\boldsymbol{\vartheta}}

\def\balpha{\mbox{\boldmath{$\alpha$}}}
\def\bbeta{\mbox{\boldmath{$\beta$}}}
\def\bdelta{\mbox{\boldmath{$\delta$}}}
\def\bgamma{\mbox{\boldmath{$\gamma$}}}
\def\blambda{\mbox{\boldmath{$\lambda$}}}
\def\bsigma{\mbox{\boldmath{$\sigma$}}}
\def\btheta{\mbox{\boldmath{$\theta$}}}
\def\bomega{\mbox{\boldmath{$\omega$}}}
\def\bxi{\mbox{\boldmath{$\xi$}}}

\def\bPsi{\mbox{\boldmath $\Psi $}}
\def\bone{\mbox{\bf 1}}
\def\bzero{\boldsymbol{ 0}}

\def\WB{{\bf WB}}

\def\A{{\bf A}}
\def\B{{\bf B}}
\def\C{{\bf C}}
\def\D{{\bf D}}
\def\E{{\bf E}}
\def\F{{\bf F}}
\def\G{{\bf G}}
\def\H{{\bf H}}
\def\I{{\bf I}}
\def\J{{\bf J}}
\def\K{{\bf K}}
\def\L{{\bf L}}
\def\M{{\bf M}}
\def\N{{\bf N}}
\def\O{{\bf O}}
\def\P{{\bf P}}
\def\Q{{\bf Q}}
\def\R{{\bf R}}
\def\S{{\bf S}}
\def\T{{\bf T}}
\def\U{{\bf U}}
\def\V{{\bf V}}
\def\W{{\bf W}}
\def\X{{\bf X}}
\def\Y{{\bf Y}}
\def\Z{{\bf Z}}

\def\a{{\bf a}}
\def\b{{\bf b}}
\def\c{{\bf c}}
\def\d{{\bf d}}

\def\f{{\bf f}}
\def\g{{\bf g}}
\def\h{{\bf h}}
\def\i{{\bf i}}
\def\j{{\bf j}}
\def\k{{\bf k}}
\def\l{{\bf l}}
\def\m{{\bf m}}
\def\n{{\bf n}}
\def\o{{\bf o}}
\def\p{{\bf p}}
\def\q{{\bf q}}
\def\r{{\bf r}}
\def\s{{\bf s}}
\def\t{{\bf t}}
\def\u{{\bf u}}
\def\v{{\bf v}}
\def\w{{\bf w}}
\def\x{{\bf x}}
\def\y{{\bf y}}
\def\z{{\bf z}}

\def\vbphi{\vec{\mbox{\boldmath $\phi$}}}
\def\vbtau{\vec{\mbox{\boldmath $\tau$}}}
\def\vbtheta{\vec{\mbox{\boldmath $\theta$}}}
\def\vI{\vec{\bf I}}
\def\vR{\vec{\bf R}}
\def\vV{\vec{\bf V}}

\def\mvec{\vec{m}}
\def\fvec{\vec{f}}
\def\appfvec{\vec{f}_k}
\def\avec{\vec{a}}
\def\bvec{\vec{b}}
\def\evec{\vec{e}}
\def\uvec{\vec{u}}
\def\xvec{\vec{x}}
\def\wvec{\vec{w}}
\def\gradvec{\vec{\nabla}}

\def\aM{\mbox{\bf a}_M}
\def\aS{\mbox{\bf a}_S}
\def\aO{\mbox{\bf a}_O}
\def\aL{\mbox{\bf a}_L}
\def\aP{\mbox{\bf a}_P}
\def\ai{\mbox{\bf a}_i}
\def\aj{\mbox{\bf a}_j}
\def\an{\mbox{\bf a}_n}
\def\a1{\mbox{\bf a}_1}
\def\a2{\mbox{\bf a}_2}
\def\a3{\mbox{\bf a}_3}
\def\a4{\mbox{\bf a}_4}

\def\sx{\mbox{\scriptsize\bf x}}
\def\st{\mbox{\scriptsize\bf t}}
\def\ss{\mbox{\scriptsize\bf s}}
\def\cR{{\cal R}}
\def\calD{{\cal D}}
\def\calS{{\cal S}}

\def\sigmae{\sigma}
\def\sigmam{\sigma}

\def\balpha{\boldsymbol{\alpha}}
\def\bbeta{\boldsymbol{\beta}}
\def\bdelta{\boldsymbol{\delta}}
\def\bgamma{\boldsymbol{\gamma}}
\def\blambda{\boldsymbol{\lambda}}
\def\bsigma{\boldsymbol{\sigma}}
\def\btheta{\boldsymbol{\theta}}
\def\bomega{\boldsymbol{\omega}}
\def\bxi{\boldsymbol{\xi}}

 \newcommand{\ie}{\emph{i.e.\;}}
 \newcommand{\eg}{\emph{e.g.\;}}
 \newcommand{\etal}{\emph{et al.\;}}
 \newcommand{\etc}{\emph{etc.\;}}

\mainmatter

\title{Random Decision Stumps for  \\ Kernel Learning and Efficient SVM}

\titlerunning{Random Decision Stumps for Kernel Learning and Efficient SVM}

\author{Gemma Roig *
\hspace{1cm}
Xavier Boix *
\hspace{1cm}
Luc Van Gool
}

\authorrunning{Gemma Roig, Xavier Boix, Luc Van Gool}

\institute{Computer Vision Lab, ETH Zurich, Switzerland\\
{\tt\small \{boxavier,gemmar,vangool\}@vision.ee.ethz.ch}\\
\small * \emph{Both first authors contributed equally.}}

\maketitle

\begin{abstract}
We propose to learn the kernel of an SVM as the weighted sum of a large number of simple, randomized binary stumps. 
Each stump takes one of the extracted features as input. This leads to an efficient and very fast SVM, while also alleviating the task of kernel selection. 
We demonstrate the capabilities of our kernel on $6$ standard vision benchmarks, in which we combine several common image descriptors, namely histograms (Flowers17 and Daimler), 
attribute-like descriptors (UCI, OSR, and a-VOC08), and Sparse Quantization (ImageNet). Results show that our kernel learning adapts well to these different feature types, 
achieving the performance of kernels specifically tuned for each, and with an evaluation cost similar to that of efficient SVM  methods. 
\end{abstract}
\setlength{\textfloatsep}{0.3cm}%

\section{Introduction}

The success of Support Vector Machines (SVMs), \eg in object recognition, stems from 
their well-studied optimization and their use of kernels to solve non-linear classification 
problems.
Designing the right kernel in combination with appropriate image descriptors is crucial. 
Their joint design leads to a \emph{chicken-and-egg} problem in that the right kernel 
depends on the image descriptors, while the image descriptors are designed for familiar 
kernels.

Multiple Kernel Learning (MKL)~\cite{Bach04} eases kernel selection by automatically 
learning it as a combination of given base kernels. Although MKL has been successful in
various vision tasks (\eg~\cite{Orabona12,Vedaldi09}), it might lead to complex and 
inefficient kernels. Recently, Bazavan~\etal~\cite{Bazavan12} introduced an approach to 
MKL that avoids the explicit computation of the kernel. It efficiently approximates 
the non-linear mapping of the hand-selected kernels~\cite{li2012chebyshev,Rahimi07,Vedaldi11}, thus delivering 
impressive speed-ups.

We propose another way around kernel learning that also allows for efficient SVMs.
Instead of combining fixed base kernels, we investigate the use of random binary 
mappings (BMs). We coin our approach Multiple Binary Kernel Learning (MBKL). 
Given that other methods based on binary decisions such as Random 
Forests~\cite{Breiman01} and Boosting decision stumps~\cite{Vezhnevets05} have not 
performed equally well on image classification benchmarks as kernel SVMs, it is all
the more important that we will show MBKL does. Not only does MBKL alleviate the 
task of selecting the right kernel, but the resulting kernel is very efficient to
compute and can scale to large datasets.

At the end of the paper, we report on MBKL results for $6$ computer vision benchmarks, 
in which we combine several common image descriptors. These descriptors are histogram-based 
(Flowers17~\cite{Nilsback06} and Daimler~\cite{Munder06}), attribute-based 
(OSR~\cite{Parikh11}, a-PASCAL VOC08 detection~\cite{aPascal}, and UCI~\cite{UCI2010}), 
and Sparse Quantization~\cite{Boix12SQ} (ImageNet~\cite{imagenet_cvpr09}).
We demonstrate for the first time that a classifier based on BMs can achieve performances
comparable to those of the hand-selected kernels for each specific descriptor. Moreover, 
it is as fast as the fastest kernel approximations, but without the need of interactively 
selecting the kernel.


\section{Efficient SVM and Kernel Learning}

In this section, we revisit the SVM literature, with special emphasis on efficient
and scalable kernel learning for object recognition.

\paragraph{\bf Efficient SVM.} 
We use $(\w,b)$ to denote the parameters of the SVM model, and $\phi(\x)$ for the 
non-linear mapping to a higher-dimensional space. The classification score for a 
feature vector $\x$ then is $\w^T \phi(\x)+b$. The SVM aims at minimizing the hinge 
loss. For the SVM implementation, one typically applies the kernel trick: with 
Lagrange multipliers $\{\alpha_i\}$ the classification score becomes 
\begin{eqnarray}
 \w^T \phi(\x) +b = &\sum_i \alpha_i y_i\left(\phi(\x)^T\phi(\x_i)\right) + b \nonumber \\
                  = &\sum_i \alpha_i y_i K(\x,\x_i) + b
\label{eqSVM}
\end{eqnarray}
where $K(\x,\x_i): \mathbb{R}^n \times \mathbb{R}^n \rightarrow \mathbb{R}$. The 
optimal multipliers $\alpha_i$ tend to be sparse and select relatively few `support 
vectors' from the many training samples. The kernel trick bypasses the computation 
of the non-linear mapping by directly computing the inner products $K(\x,\x_i)$.

The strength of SVMs is that they yield max-margin classifiers. At test time, the
computational cost is the number of support vectors times the cost of computing the 
kernel. The problem is that the latter may be quite expensive. Also, during training,
the complexity of computing the kernel matrix grows quadratically with the number of 
training images, which renders it intractable for large datasets.

Several authors have tried to speed up kernel-based classification. Ideas include 
limiting the number of support vectors~\cite{Burges97,Keerthi06} or creating low-rank 
approximations of the kernel matrix~\cite{Fine01LowRank}. These methods are effective, 
but do not scale well to large datasets because they require the kernel distances to 
the training set. Rahimi and Recht introduced Random Fourier Features~\cite{Rahimi07},
thereby circumventing the approximation of the explicit feature map, $\phi(\x)$. Such
techniques have been explored further for kernels used with common image descriptors, 
such as $\chi^2$ and RB-$\chi^2$ kernels~\cite{li2012chebyshev,Vedaldi11} or the 
intersection kernel~\cite{Maji12,WuHIK}. Other approaches use kernel PCA to linearize 
the image descriptors~\cite{Perronin2010Embedding} or sparse feature embeddings~\cite{vedaldi12sparse}. 
Recently, Wu~\cite{Wu12PmSVM} introduced the power mean kernel, which generalizes the 
intersection and $\chi^2$ kernels, among others, and achieves a remarkably efficient, 
scalable SVM optimization.

These methods approximate specific families of kernels. Our aim is to \emph{learn} a fast 
kernel, which eases kernel selection, rather than approximating predefined kernels.

\paragraph{\bf Multiple Kernel Learning (MKL).}
MKL~\cite{Bach04} aims at jointly learning the SVM and a linear combination of given 
base kernels. The hope is that such committee of base kernels is a more powerful kernel.
Denoting the base kernels as $\hat{K}$, the final kernel $K$ takes the form 
 \begin{align}
K(\x,\x_i) = \sum_k \theta_k \hat{K}_k(\x,\x_i),
 \label{eqAtomicKernel}
 \end{align}
where the  weights $\theta_k\in \mathbb{R}_{+}$ can be discriminatively learned. 
There 
have also been some approaches to find non-linear combinations of kernels, \eg.~\cite{Cortes09,Varma09}, that we do not further consider here. 

In recent years, many advances have been made to improve the efficiency of MKL, and 
various optimization techniques have been introduced, \eg semi-definite programming~\cite{Lanckriet04}, SMO~\cite{Bach04,Vishwanathan10}, semi-infinite linear programming 
\cite{Sonnenburg06} and gradient-based methods~\cite{Rakotomamonjy08,Varma09}. 
Yet, scalability to large datasets remains an issue, as these methods explicitly compute 
the base kernel matrices. Therefore, Bazavan~\etal~\cite{Bazavan12} exploit Random 
Fourier Features, which approximate the non-linear mapping of the kernel, and allow to 
scale to large datasets. 

Our approach is related to the latter in that it also aims at efficient and scalable 
kernel learning. Yet, MBKL's base kernels are not hand-selected. Instead of approximating 
a distance coming with a pre-selected kernel, we explore the use of random BMs to 
{\em learn} a distance for classification. Indeed, in large-scale image retrieval, 
there is an increasing body of evidence that suggests that BMs are effective to evaluate 
distances,~\eg~\cite{Raginsky09,Torralba07,Wang10}. 

In the next section, we introduce our formulation for kernel learning built from BMs.
This, in turn, will yield a kernel very efficient to learn and to evaluate ~(Section~\ref{secCompComplex}). 
Moreover, the kernel will adapt to most image descriptors, since it is learned from the 
input data~(Section~\ref{seqLearn}).


\section{Multiple Binary Kernel Learning}
\label{sec:ekl}

In this section, 
we introduce a kernel that is a linear combination of binary kernels, defined from a 
set of simple decision stumps.
 

\paragraph{\bf MKL with Binary Base Kernels.}
BMs have been used to speed-up distance computations in large-scale image retrieval,~\eg~\cite{Raginsky09,Torralba07,Wang10}. In these methods, the input feature is transformed 
into a binary vector that preserves the locality of the original feature space. In the
context of classification, we can further enforce that the kernel in an SVM separates the 
image classes well. 

We adopt the MKL formulation (see eq.~\eqref{eqAtomicKernel}) as starting point of our kernel, 
since it aims at jointly learning the classifier and the kernel distance, yet incorporate
BMs and restrict the base kernels to only take on binary values. The binary base kernels 
are defined as:
 \begin{align}
 \hat{K}_k(\x,\x_i) = \I[\sigma_k(\x) = \sigma_k(\x_i)],
 \label{defAtomicKernel}
 \end{align}
where $\I[\cdot]$ is the indicator function, which returns $1$ if the input is true and $0$ otherwise, and $\sigma_k(\x)$ is a BM of the input feature, $ \sigma_k(\x): \mathbb{R}^n 
\rightarrow \{0,1\}$. Each base kernel is built upon one single BM. The BMs need not be 
linear and can be adapted to each problem if desired. In the sequel, we explore different 
possibilities, but our kernel is not restricted to any of them. In all cases, $\sigma_k(\x)$ 
divides the feature space into two sets and the indicator function returns whether the two 
input samples fall in the same part of the feature space or not.

The final kernel, $K$, is a linear combination of the binary kernels: $K(\x,\x_i) =\sum_k 
\theta_k \I[\sigma_k(\x) = \sigma_k(\x_i)]$. Note that $K$ is not restricted to be binary, 
though the base kernels are. In the supplementary material we show that such `Multiple 
Binary Kernel' (MBK) is a valid Mercer kernel.  
An appropriate choice of the $\sigma_k(\x)$ will be 
important to arrive at good classifications. For instance, in a two class problem, the 
better the set of $\sigma_k(\x)$ separate the two classes, the better the kernel might be.

\paragraph{\bf Explicit Non-linear Mapping.}
\label{secCompComplex}
We analyze the  
 form of the non-linear mapping of MBKL. 
In the Supplementary Material, we derive the
non-linear mapping $\phi(\x)$ that induces the MBKL kernel, and it is
\begin{align}
[\sqrt{\theta_1} \sigma_1(\x), \sqrt{\theta_1} \bar{\sigma}_1(\x),  \sqrt{\theta_2}\sigma_2(\x),\sqrt{\theta_2}\bar{\sigma}_2(\x),\ldots ]^T,
\label{eqmap}
\end{align}
where $\bar{\sigma}_k(\x)$ is ${\sigma}_k(\x)$ plus the \emph{not} operation,
and the SVM parameters are 
\begin{align}
\w^T= [\sqrt{\theta_1}c^{1}_1,\; \sqrt{\theta_1}c^{0}_1, \; \sqrt{\theta_2}c^{1}_2, \; \sqrt{\theta_2}c^{0}_2, \; \ldots\; ], 
\label{eqw}
\end{align}
where $c^1_k, c^0_k \in \mathbb{R}$ are two learned constants, which correspond to the underlying parameters of the classifier.
 We can see that this mapping recovers the 
MBKL kernel in the form $\phi(\x)^T\phi(\x_i)$.
Thus, to evaluate MBKL at test time,  we do not need to evaluate the kernel 
 because we have access to the non-linear mapping $\phi$. 

\paragraph{\bf Benefits of kernel learning.}
MBKL generalizes a SVM with BMs as input features.
This can be easily seen by fixing $\btheta=\1$ in eq.~\eqref{eqmap} and~\eqref{eqw}.  
But learning $\btheta$ rather than fixing it to $\1$ has several advantages. 
Recall that the kernel distance   does not depend on the image class we are evaluating.
A BM with $\theta_k$ equal to $0$ does not contribute to the final kernel distance, and hence,
 can be discarded for {\em all} image classes. 
This is crucial to arrive at a competitive computational complexity.
 Moreover, MBKL aims at learning a kernel distance adapted to the image descriptors, that can be used
for tasks other than classification.

MBKL is not a particular instance of any of the kernels in the literature. Rather the 
opposite may be true, since most kernels can be approximated with a collection of BMs~\cite{Raginsky09}.

\section{BMs as Random Decision Stumps}
\label{secRandom}

We found that defining the $\sigma_k(\x)$ as simple random decision stumps achieves 
excellent results with the image descriptors commonly used in the literature. Decision 
stumps select a component in a feature vector and threshold it. We randomly select a 
component $i\in\mathbb{N}$ of the input feature vector, using a uniform probability 
distribution between $1$ and the feature length. Then, the BM is calculated applying 
a threshold, $\sigma_k(\x)=\I[  x_i > t]$, where $t\in \mathbb{R}$ is the threshold 
value. Again, this threshold is generated from a uniform probability distribution,
here over the interval of values observed during training for component $i$. Note that 
we generate $\sigma_k(\x)$ randomly, without using labeled data. In contrast, the   
supervised learning of the kernel and the SVM will use labeled data to appropriately
combine the BMs (Section~\ref{seqLearn}). 

We may need thousands of random BMs to 
arrive at the desired level of performance.
Since the decision 
stumps have cost $\bigO(1)$, the computational complexity of evaluating MBKL at test time grows linearly with 
the number of BMs. In the experiments we show that this is of the same order of magnitude as the 
feature length, or one order higher. This allows to achieve a competitive computational cost compared to other methods, as we report in the experimental section.

\begin{figure}[t!]
\centering
\begin{tabular}{@{\hspace{-0.8mm}}c@{\hspace{-1mm}}c@{\hspace{-1mm}}c}
\includegraphics[scale=0.267]{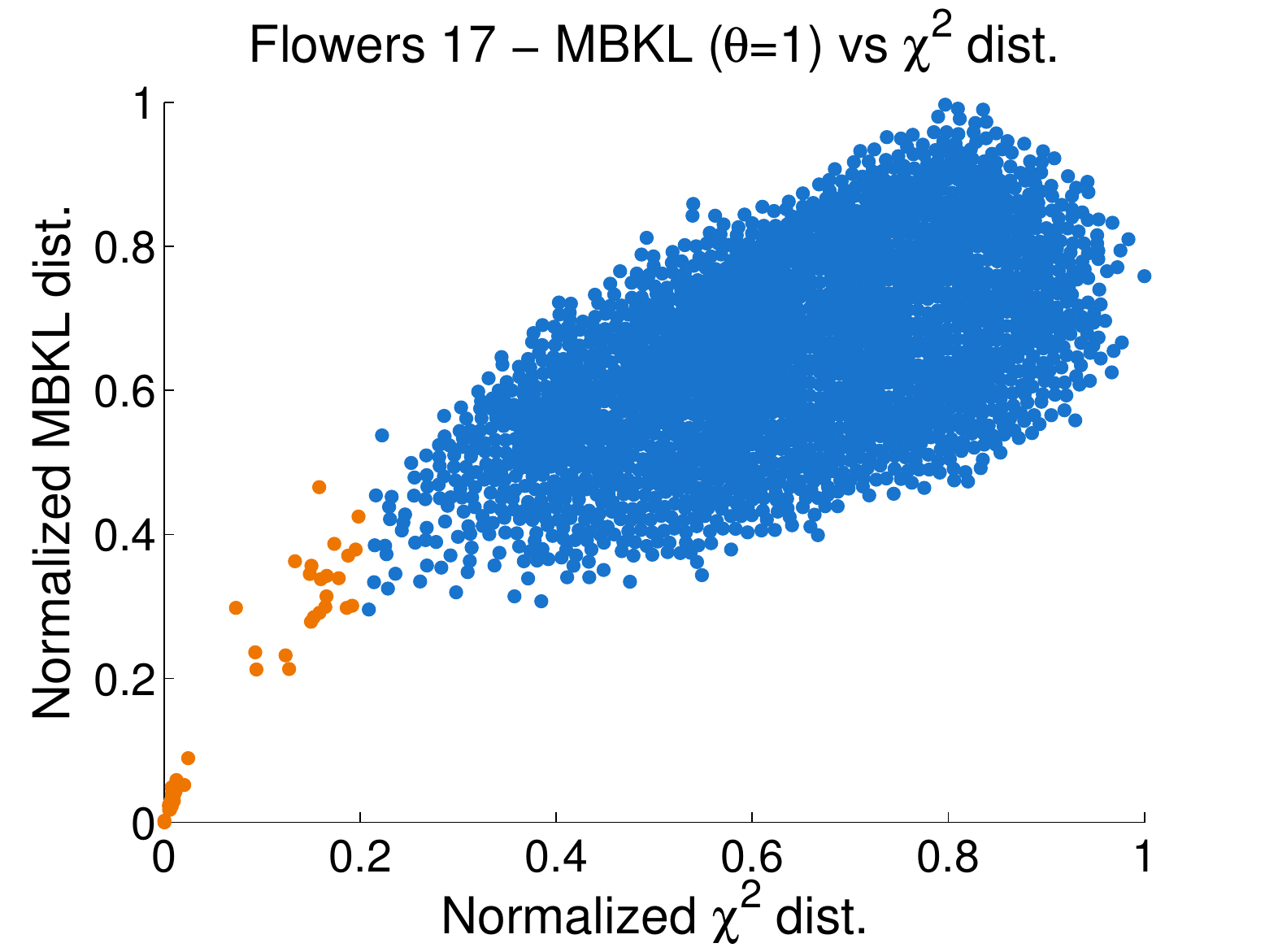} &  \includegraphics[scale=0.267]{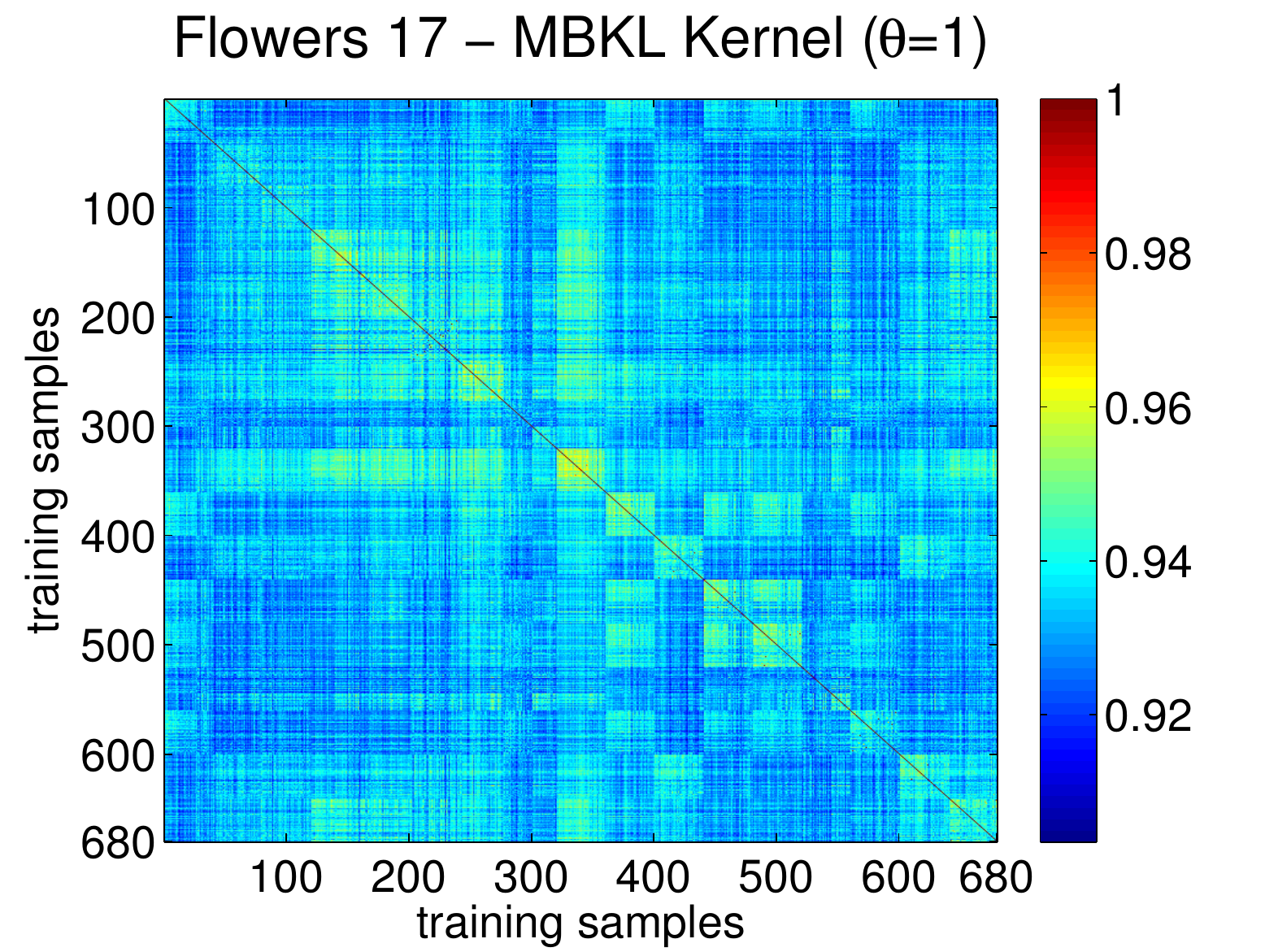} &  \includegraphics[scale=0.267]{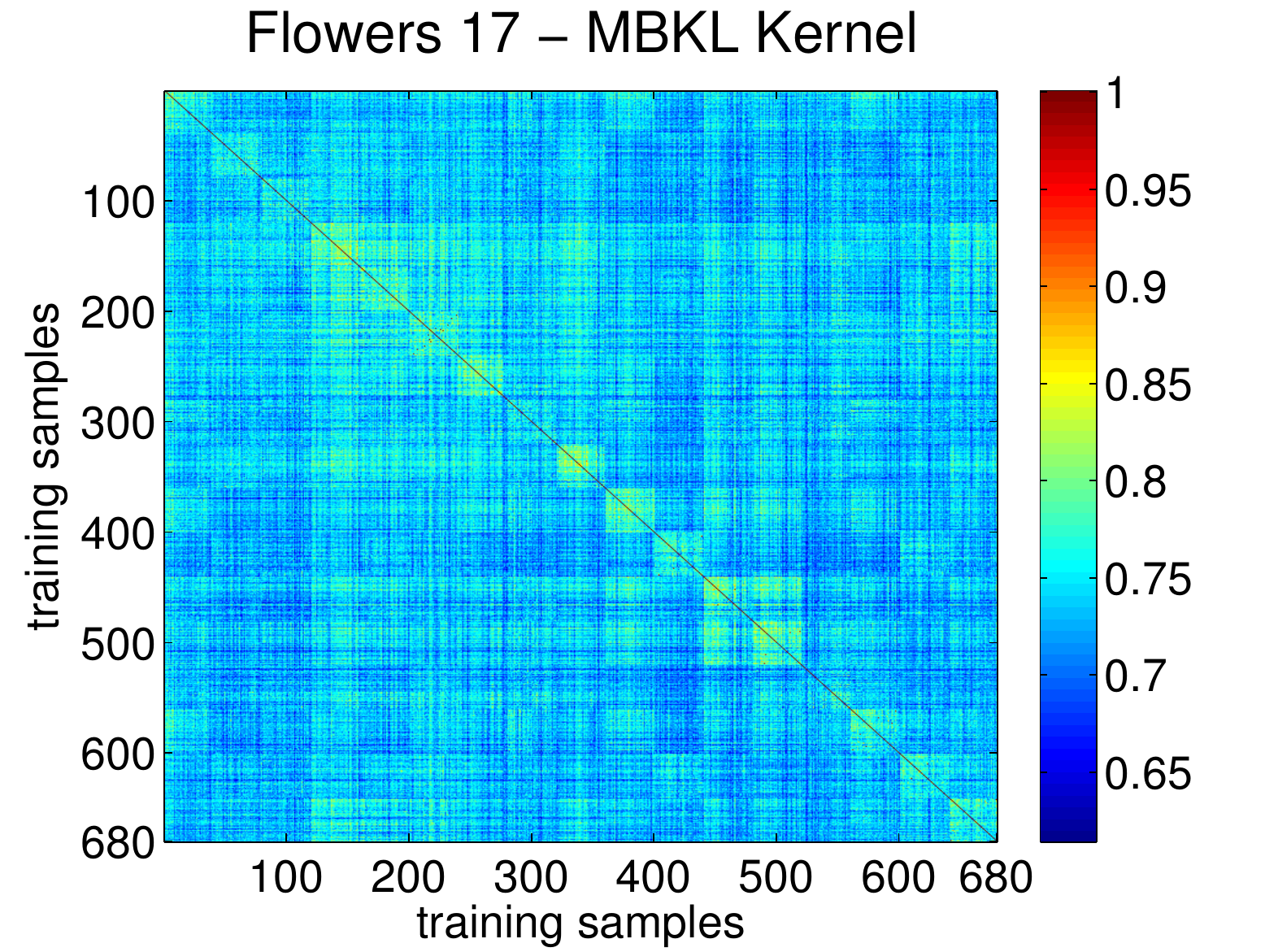} \\ 
\footnotesize{(a)} & \footnotesize{(b)} & \footnotesize{(c)}\\
\end{tabular}
\caption{ \emph{BMs for Kernel Learning}. On Flowers 17 training set: (a) Comparison between 
$\chi^2$ distance and a MBKL distance with $\btheta=\1$ (MKBL kernel is normalized with the amount of BMs). 
Each point represents the distance between two images in the training set (we indicate in orange color that the $\chi^2$ distance is lower than $0.2$). 
(b) MBKL kernel with $\btheta=\1$, and (c) MBKL kernel with the learned $\btheta$.
For (b) and (c) images are sorted with their class label, this is why some semantic clusters can be seen  around the diagonal.}
\label{figdist}
\end{figure}

Intuitively, random decision stumps may seem to quantize the image descriptor too 
crudely. That might then affect the structure of the feature space and deteriorate 
performance. Yet,  in classification, decision stumps are known to allow for good 
generalization~\cite{Amit97,Rahimi08,Rahimi08b}. 
As an illustration, Fig.~\ref{figdist}a compares the $\chi^2$ distance  
and MBKL with decision stumps. We use the experimental setup of Flowers17 (see Section~\ref{sec:experiments}), for which $\chi^2$ is the best performing kernel, but the other 
kernel distances and datasets in the paper yield the same conclusions. MBKL uses 
$30,000$ random decision stumps and for the time being we simply put $\btheta=\1$, 
\ie all $\theta_k = 1$. We can see that the distances produced by both methods are
highly correlated. The decision stumps do change the structure of the feature space, 
but keep it largely intact. Since the kernel distance is parametrized through $\btheta$, 
that modulates the contribution of each BM, MBKL can further adjust the kernel distance 
to the SVM objective. 
Fig.~\ref{figdist}b shows the final MBKL kernel, and Fig.~\ref{figdist}c the MBKL  $\btheta$-adjusted kernel, as 
learned in Section~\ref{seqLearn}. Observe that the high kernel values
between images of different classes in the non-learned kernel,  are smoothed out  in the learned kernel.

\small
\normalsize  

Also, note that using decision stumps with MBKL differs substantially from boosting decision stumps. 
Apart from the differences in the loss function, 
boosting optimizes the parameters of the BMs individually, using labeled data, and progressively adds them to the final classifier. 
 MBKL generates all the BMs all at once, with random parameters and without using labeled data.


 \begin{algorithm}[t!]
 \caption{Multiple Binary Kernel Learning}
 \label{algLearning}
\normalsize
 \KwIn{$(\x_i,y_i)$, \ \ $\forall i$}
 \KwOut{$\bsigma(\x)$, $\btheta$, $\w$  }
 $\{\sigma_k\} = $ Generate Random Tests; \\
 \emph{Step 0:} $\{c_k^1,c_k^0\} =$ Initial Guess $\left(\{\sigma_k(\x)\},\y\right);$ \\
 \emph{Step 1:} $\btheta=$ SVM$_{l_1}$ $\left(\{c_k^1,c_k^0\}, \y\right);$\\
 $\bsigma =$  Select $\{\theta_k\} > 0;$\\
\emph{Step 2:} $\{c_k^1,c_k^0\}=$ SVM $\left(\phi(\x),\y\right);$\\
 \end{algorithm}

\section{Efficient Two-stage Learning}
\label{seqLearn}
In this section, we introduce the formulation for learning the kernel and the classifier parameters, 
once the random BM have already been  generated.
MBKL pursues minimizing the SVM objective.
Rather than jointly optimizing the kernel and the classifier -- which may be not feasible in practice for thousands of binary kernels --  
we decompose the learning in two stages to make it tractable.
All stages optimize the same SVM objective, but either $\btheta$ or the classifier parameters  are kept fixed. 
Firstly, we fix the classifier parameters to  an initial guess, and we learn the kernel, $\btheta$.  Secondly,  
the classifier is trained with the learned kernel.
We could extend this algorithm to iteratively re-learn $\btheta$ and the classifier parameters, but this would obviously raise the computational  cost while we did not observe any increase in performance. 
Also, note that we do not re-sample new BMs after discarding some.

Next we describe the two steps of the learning in more detail. Before the actual optimization 
starts, we have to initialize the classifier, which we describe as the prior Step $0$. Step $1$ then 
learns the kernel parameters, after which Step $2$ learns the actual classifier. We show 
that the most complex optimizations can be solved with off-the-shelf SVM solvers in the primal form. 
We summarize all steps of the learning in Algorithm~\ref{algLearning}.

\paragraph{\bf Step $0$: Efficient Initial Guess of the Classifier.}
Recall that each binary kernel has two parameters associated: $c_k^1,c_k^0$ (eq.~\eqref{eqmap}, \eqref{eqw}). 
In order to efficiently get an initial guess of these parameters,  we learn each pair of parameters  individually, without taking into account the other kernels.
The downside is that this form of learning is rather myopic, blind as it is to the information coming from the other kernels. 
However, this is alleviated by the global learning of the kernel weights $\btheta$ and the SVM classifier in Steps 1 and 2 of the algorithm.

For the initial guess of $(c_k^1,c_k^0)$ we also use the SVM objective, but we formulate it for each kernel individually.
The classifier and the non-linear mapping of a single kernel becomes 
\begin{align}
\phi_k(\x)^T=\sqrt{\theta_k}(\sigma_k(\x),\bar{\sigma}_k(\x)), \; \; \w_k^T=\sqrt{\theta_k}(c_k^1,  c_k^0), 
\end{align}
and we place them in an SVM objective function.
We can ignore the dependence on $\sqrt{\theta_k}$ because 
it only scales the classification score, and can be compensated by $\w_k$. Thus, 
\begin{align}
\phi_k(\x)^T=(\sigma_k(\x),\bar{\sigma}_k(\x)), \; \; \w_k^T=(c_k^1,  c_k^0). 
\end{align}
Interestingly, because when $\sigma_k(\x)$ is $1$ then $\bar{\sigma}_k(\x)$ is $0$, and v.v., $\w_k^T \phi_k(\x)$  can  only take two values, 
\ie, either $[1, 0][c_k^1,  c_k^0]^T=c_k^1$ or $[0, 1][c_k^1,  c_k^0]^T=c_k^0$.
As a consequence, we can show  that  
when we optimize  $\w_k^T$  with a linear SVM with input features $(\sigma_k(\x),\bar{\sigma}_k(\x))$, then, $\w^T=(a,  -a)$,
where $a\in \mathbb{R}$ (see  Supplementary Material).

This shows that the SVM classifier for one binary kernel only requires learning a single parameter, $a$. 
Further, introducing this result into  $\w_k^T\phi_k(\x)$ yields 
\begin{align}
\w_k^T\phi_k(\x)=(a \I [ \sigma_k(\x)=1 ] -a  \I [\sigma_k(\x)=0 ] ). 
\label{eqAA}
\end{align}
If we let $a=sign(a)|a|$, and discard $|a|$ because it is only a scale factor that can be later absorbed by $\theta_k$ (if $a\neq 0$), 
we obtain that $[c_k^1,  c_k^0]$ is determined by $sign(a)$ when $a\neq 0$. Thus, 
\begin{align}
\w_k^T\phi_k(\x) = sign(a) \I [\sigma_k(\x)=1 ] - sign(a) \I [\sigma_k(\x) = 0], 
\end{align}
when $a\neq 0$.
 Using  the proportion of samples that responded $\sigma_k(\x)=1$, we can determine the sign of $a$, and when $a$ is $0$. 
Let $t_p$  be the number of samples of the class we are learning the classifier (positive sample), and $t_n$ 
the number of samples for the rest of classes (negative).  
Let $p$ and $n$ be how many samples of each class have test value $\sigma_k(\x)=1$.
Table~\ref{tableSimpleSVM}  shows in which cases  $a=0$ or, otherwise $sign(a)$.
In case  $a=0$  we discard it, since in Eq.~\eqref{eqAA} 
$a=0$  can not contribute in any way
to the final classification score.
These rules can be  deduced by
fulfilling the max-margin of the SVM objective.
For of multi-class problem,
we do the one-vs-rest strategy, and initialize the parameters $(c_k^1,c_k^0)$
independently for each classifier.

\begin{table}[t]
\centering
\begin{tabular}{|c|c|c|}
\hline
 & $\mathbf{p/t_p\geq 0.5}$ & $\mathbf{p/t_p<0.5}$ \\ \hline
 $\mathbf{n/t_n\geq 0.5}$ &	$a=0$  &  $sign(a)=+1$ \\ \hline
 $\mathbf{n/t_n< 0.5}$ &  $sign(a)=-1$ & $a=0$ \\ \hline
\end{tabular}
\caption{ \emph{Learning of $\w_k^T=[c_k^1,  c_k^0]$ through parameter $a$. } $t_p$ and $t_n$ are the number of positive and negative samples, respectively, 
and $p$ and $n$ how many samples of each class have  test value $\sigma_k(\x)=1$.}
\label{tableSimpleSVM}
\end{table}

\paragraph{\bf Step $1$: Learning the Kernel Parameters.}
We fix  $(c_k^1,c_k^0)$ using the initial guess previously learned and we learn the $\btheta$ that minimizes 
the SVM loss.
 In order to write the SVM loss directly depending on $\btheta$,
we reorder $\w^T \bphi(\x)$. From eq.~\eqref{eqmap} and~\eqref{eqw}, we obtain $\w^T \bphi(\x) =  \btheta ^T \s(\x)$, where 
  $\s(\x) = [s_1(\x),\dots,s_k(\x),\dots]$, and $s_k(\x)=c_k^1\sigma_k(\x) +c_k^0\bar{\sigma}_k(\x)$. Note that $s_k(\x)$ is
known because it only depends on the already guessed $c_k^1,c_k^0$, and $\sigma_k(\x)$.
With some algebra, the SVM objective that we are pursuing becomes
\begin{align}
 \min_{\btheta,\bxi }  \|\btheta \|_1  + C \|\bxi\|_1, 
 \;\mbox{s.t.}  \forall i: y_i \btheta ^T \s(\x) +b \geq 1 - \xi_{i}. \label{eqSVMTHETA} 
\end{align}
Observe that the regularizer becomes $\|\w \|^2_2=\|\btheta \|_1 $ since $\{c_k^1,c_k^0\}$ is either
$1$ or $-1$, and can be discarded because of the square. Thus, we learn $\btheta$ with a 2-class linear $\ell_1$-SVM
where the input features are $\s(\x)$.
Since the off-the-shelf SVM optimizers do not constrain $\theta_k\geq 0$, 
it might happen that for some kernels this is not fulfilled.
In that case, we directly set the $\theta_k$ that are not positive to $0$. 
Note that this is necessary to yield a valid Mercer kernel.

For a  multi-class problem,  
the kernel parameters, \ie $\{\sigma_k(\x)\}$ and $\btheta$, are the same for all classes, whereas there is a specific set 
 $\{(c_{k}^0,c_{k}^1)\}$ for each class, denoted as $\{(c_{k,y}^0,c_{k,y}^1)\}$.
We follow the same learning strategy, 
(\ie we optimize $c_{k,y}^0,c_{k,y}^1$ according to Table~\ref{tableSimpleSVM}), but using the multi-class heuristic of one-vs-all. 
Let $s_y(\x_i)$ be the responses corresponding to class $y$. 
Because $\btheta$ is the same for all classes (the kernel does not change with  the class we are evaluating),
 the SVM in  eq.~\eqref{eqSVMTHETA} is still a \emph{two-class} SVM in which
we take as positive samples the $s_y(\x_i)$ evaluated for the true class, (\ie when $y=y_i$), 
and the others as negative. This may yield a large negative training set, but we found that in  practice it suffices to use a reduced subset of examples.
In practice, we generate the subset of negative samples by randomly extracting examples whose object class is different from the target class, and taking into account 
that the amount of examples per object class is balanced. The number of samples for each dataset is detailed in the experiments section.

\paragraph{\bf Step $2$: Learning the Classifier.}
Finally, we use the learned kernel to train a standard SVM classifier,
thus replacing the initial guess of
the classifier. 
We discard the
BMs with $\theta_k=0$ before learning the SVM, because they do not contribute to the final kernel. 
From eq.~\eqref{eqmap} and~\eqref{eqw},
we can deduce that optimizing the SVM objective with $\{c_k^1,c_k^0\}$ as the only remaining parameters can be done with an SVM in the primal form with
$\phi(\x)=[\theta_1\sigma_1(\x), \theta_1\bar{\sigma}_1(\x),  \theta_2\sigma_2(\x),
\theta_2\bar{\sigma}_2(\x),\ldots ]$ and
$\w^T= [c^{1}_1,\; c^{0}_1, \; c^{1}_2, \; c^{0}_2, \; \ldots\; ]$.

\paragraph{\bf Computational Cost and Scalability.} 
We can see by analyzing all the steps of the algorithm that it scales linearly to the number of training data.
Step~$0$ only requires evaluating $\sigma_k(\x)$ for the training set, and use the 
simple rules in Table~\ref{tableSimpleSVM}.
Besides, since the $\{c_k^1,c_k^0\}$ are learned independently, this can be parallelized.
Step~$1$ and~$2$ are optimized with SVMs  in the primal form.
Note that for most practical cases, the computational cost of Step~$1$ may be the bottleneck of the algorithm.
The feature length of Step~$1$ is equal to the initial number of BMs, which is larger than  
the number of BM with $\theta_k\neq 0$, used in Step~$2$.

Moreover, all steps of the learning algorithm also scale linearly to the number of classes.
Note that Step~$0$ and~$2$ can be solved with the one-vs-all strategy, and 
Step~$1$ is always a two-class SVM.


\providecommand{\e}[1]{\ensuremath{\times 10^{#1}}}

\section{Experiments}
\label{sec:experiments}
In this section we report the experimental results on $6$ benchmarks, in order to evaluate MBKL in the context of a variety of vision tasks and image descriptors.
 After introducing the most relevant implementation details, 
 we discuss the results.

\subsection{Experimental Setup}
All the experiments are run on $4$ CPUs Intel i7@$3.06$GHz. 
We chose the $C$ parameter of the SVM among $\{0.01, 0.1 , 1, 10 , 100, 1000\}$ 
by cross-validation on the training set, and we fix it for computing the times. We use liblinear~\cite{LIBLINEAR} library when using linear SVM, and libsvm~\cite{LIBSVM} library
 when using a kernel.

Table~\ref{tdatasets} summarizes the datasets used, as well as their characteristics and the features used.  
In case the features are attributes, we normalize them with a logistic function to lie in the interval $[0,1]$. We use this 
normalization for all methods. 
For each dataset, the standard evaluation procedures described in the literature are used. Further details are provided in the Supplementary Material.
We evaluate MBKL's efficiency for all the datasets except UCI, for which the computational cost is very little for all methods.

 \begin{table}[t!]
\centering
\begin{tabular}{|@{\hspace{1mm}}l@{}@{\hspace{3mm}}|@{\hspace{3mm}}c@{\hspace{3mm}}c@{\hspace{3mm}}c@{\hspace{3mm}}c@{\hspace{3mm}}c@{\hspace{3mm}}c@{\hspace{3mm}}c@{\hspace{3mm}}|}

\hline
\emph{Dataset}&\bf{Daimler }	& \bf{Liver} 	& \bf{Sonar}  		& \bf{Flowers17}	&\bf{OSR}				& \bf{a-VOC08}		&\bf{ImgNet}\\
\hline
\hline
\emph{\# Classes}&$2$&  $2$ 	&$2$			& $17$ 			&$8$ 						&$20$ 			& $909$\\
\emph{\# Im. Train}	&$19,600$& $276$ 	&$166$			& $1,020$ 		&$240$ 	 		&$6,340$ 	&   $1e{6}$\\
\emph{\# Im. Test}		& $340$ 		&$2,448$ 		& $69$ 	&$32$	&$9,800$ 		&$6,355$ 	&  $5e{4}$\\
\emph{Descr.}& HoG	&  Attr 	&Attr			& BoWs 			& HoG+Attr				& HoG+Attr 		&  S.Quant.\\
\emph{Feat. Len.}& $558$& $6$ 	 &$60$			& $31e{3}$ 	&$518$ 						&$9,752$  		& $21e{3}$ \\
\hline
 \end{tabular}
   \caption{ \emph{Summary of the Datasets.} The number of images for training and testing are reported
for $1$ split.}
   \label{tdatasets}
\end{table}

\vspace*{-0.1cm}
\paragraph{$-$ Daimler~\cite{Munder06} (Pedestrian detection).} 
This is a two-class benchmark, consisting of $5$ disjoints sets, 
each of them containing $4,800$ pedestrian samples and $5,000$ non-pedestrian examples.
We use $3$ splits for training and $2$ for testing. Testing is done on the two other sets separately, 
yielding a total of 6 testing results. The HoG descriptor is used. 

\vspace*{-0.1cm}
\paragraph{$-$ UCI~\cite{UCI2010} (Object Recognition).} 
We report results on two-class problems, namely Liver  and Sonar, using $5$ 
cross-validations. We use the attribute-based descriptors that are provided, which are of length $6$ and $60$, respectively.

\vspace*{-0.1cm}
\paragraph{$-$ Flowers17~\cite{Nilsback06} (Image classification).} 
It consists of 
17 different kinds of flowers with 80 images per class, divided in $3$ splits.
We describe the images using the features provided by~\cite{Khan09}: SIFT, opponent SIFT, WSIFT  
and color attention (CA), building a Bag-of-Words histogram, computed using spatial pyramids. 

\vspace*{-0.1cm}
\paragraph{$-$ OSR~\cite{Parikh11} (Scene recognition).} 
It contains $2,688$ images from 8 categories, of which $240$ are used for training and the rest for testing. We use the $512$-dimensional
GIST descriptor and the $6$ relative attributes provided by the authors of~\cite{Parikh11}.

\vspace*{-0.1cm}
\paragraph{$-$ a-PASCAL VOC08~\cite{aPascal} (Object detection).} 
It consists of $12,000$ images of objects divided in train and validation sets. 
The objects were cropped from the original images of VOC08. 
There are $20$ different categories, with $150$ to $1,000$ examples per class, except for people 
with $5,000$. 
The features provided with the dataset are local texture, HOG and color descriptors. For each
image also $64$ attributes are given. In~\cite{aPascal} it is reported that those attributes were obtained by asking users in Mechanical Turk for semantic attributes for
each object class in the dataset. They can be used to improve classification accuracy.
We use both the features and the attributes. 

\vspace*{-0.1cm}
\paragraph{$-$ ImageNet~\cite{imagenet_cvpr09}.}
We create a new dataset taking a subset of $1,065,687$ images. 
This subset contains images of $909$ different classes that do not 
overlap with the synset. We randomly split this subset into
a set of $50,000$ images for testing and the rest for training, maintaining the proportion of images per class.
For evaluation, we report the average classification accuracy across all classes. 
We use the Nested Sparse Quantization descriptors, provided by~\cite{Boix12SQ}, using their setup.  
We use $1024$ codebook entries with max-pooling 
in spatial pyramids of $21$ regions ($1\times 1$, $2\times 2$ and $4\times 4$). As for many state-of-the-art (s-o-a)
descriptors for image classification, better accuracy is achieved with a linear SVM than with kernel SVM.
Additionally, 
we test a second descriptor in ImageNet. We use the same setup as~\cite{Boix12SQ}
to create a standard Bag-of-Words by replacing the max-pooling by average pooling. 
This descriptor performs better with kernel SVM, but it achieves lower performance
than max-pooling in a linear SVM.

\begin{figure*}[t!]
\centering
 \begin{tabular}{c@{\hspace*{0.5cm}}c}
  \includegraphics[scale=0.32]{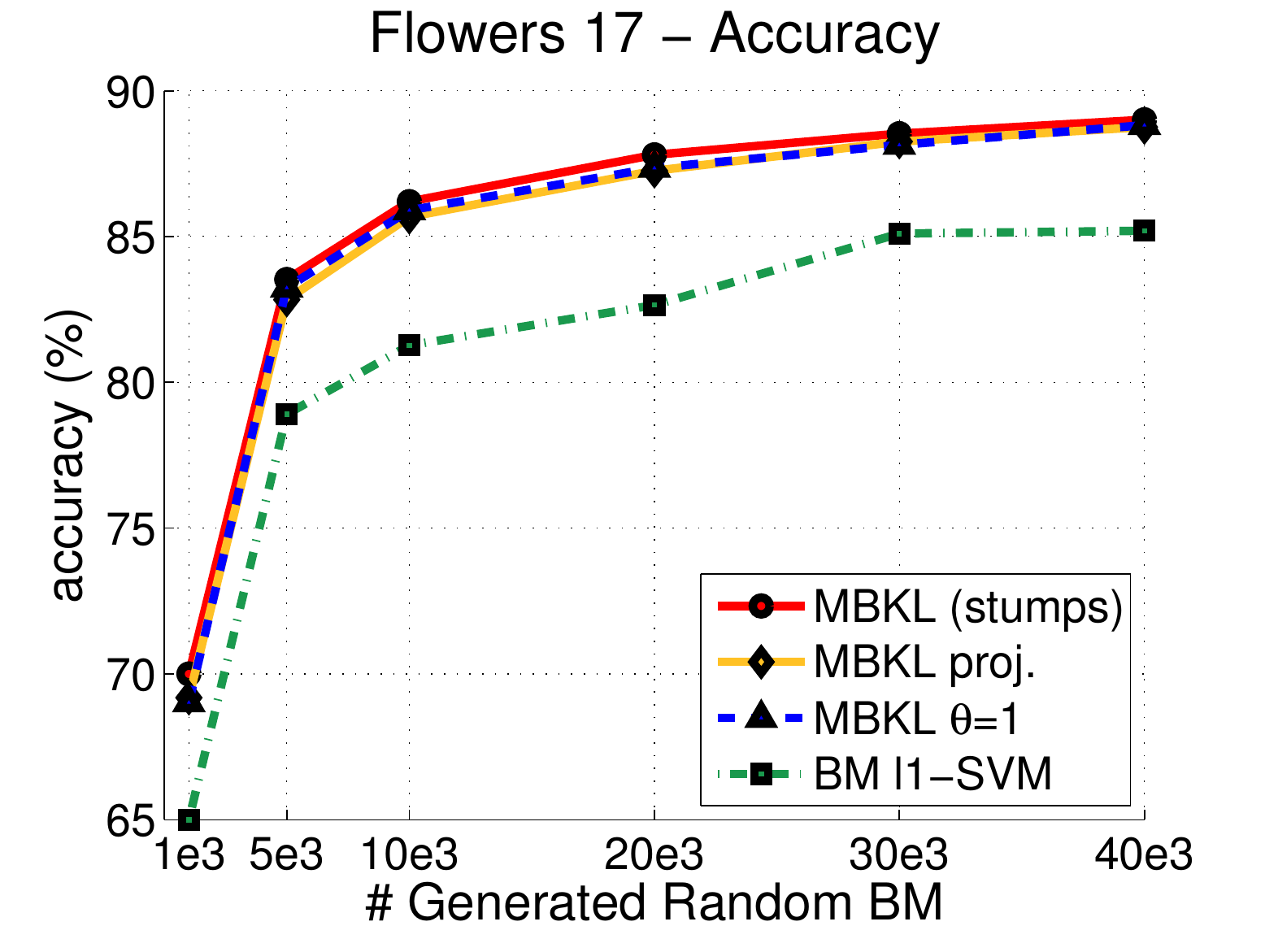} &
\includegraphics[scale=0.32]{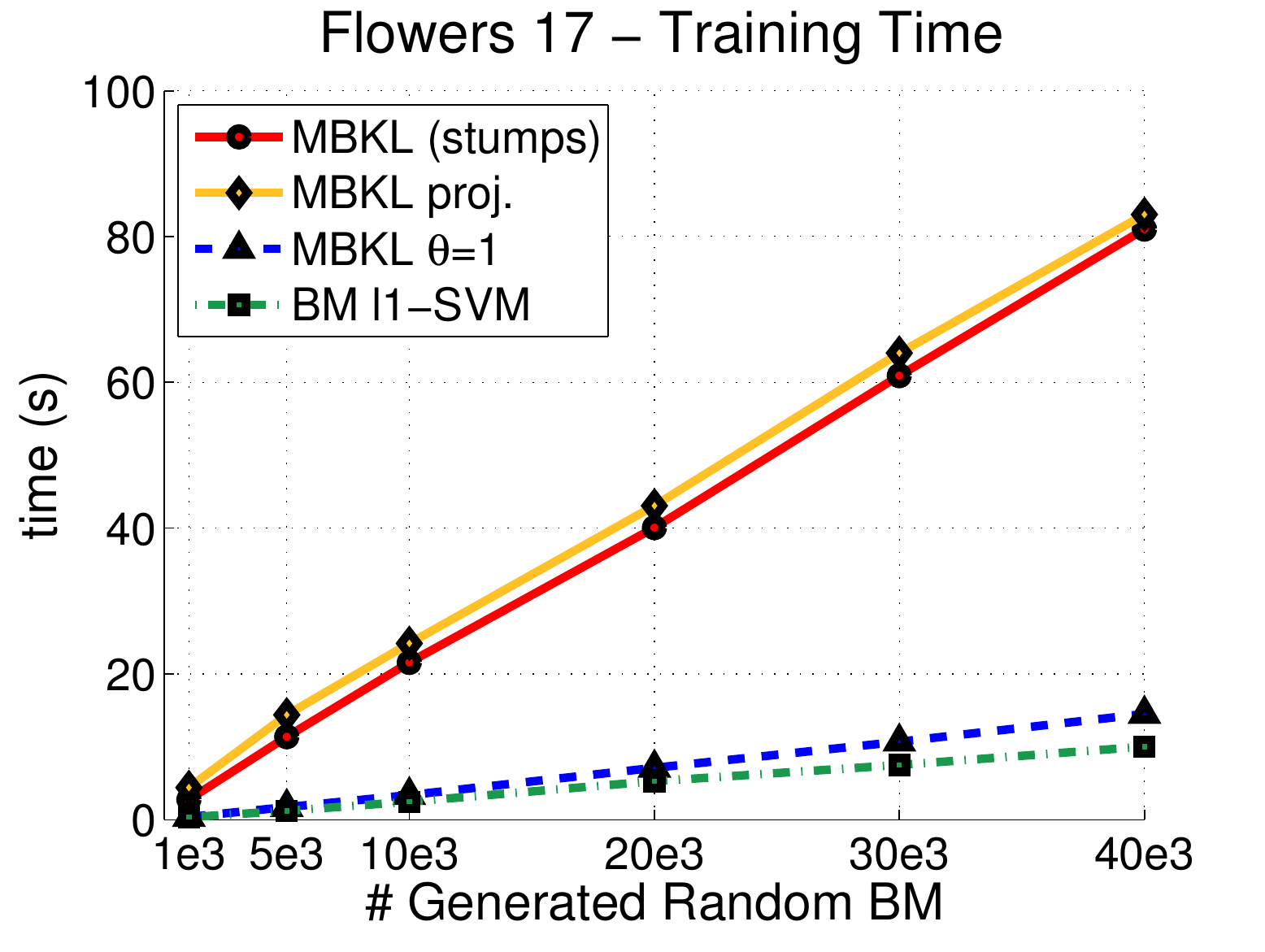}\\
(a) & (b) \\
  \includegraphics[scale=0.32]{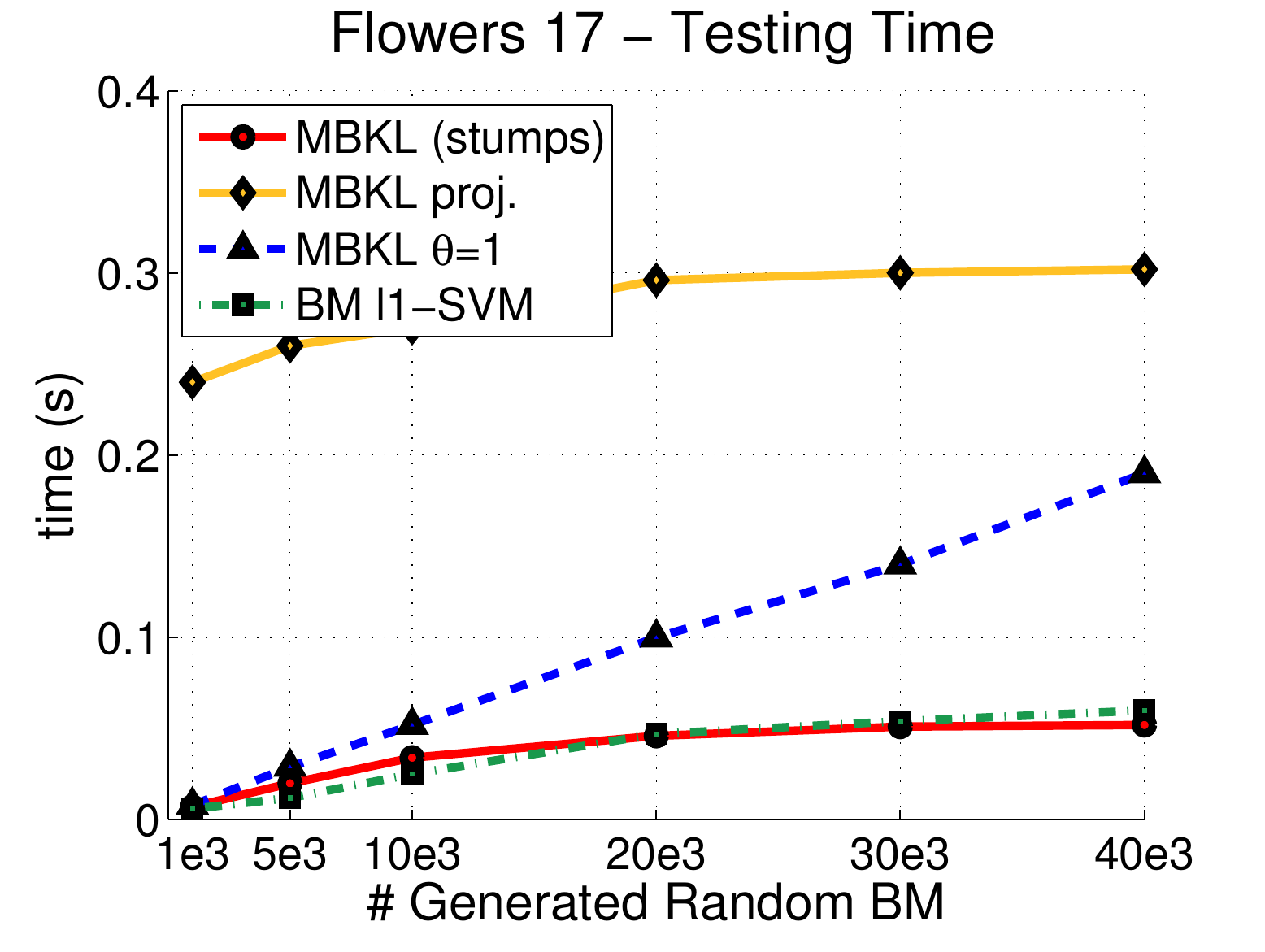}&
  \includegraphics[scale=0.32]{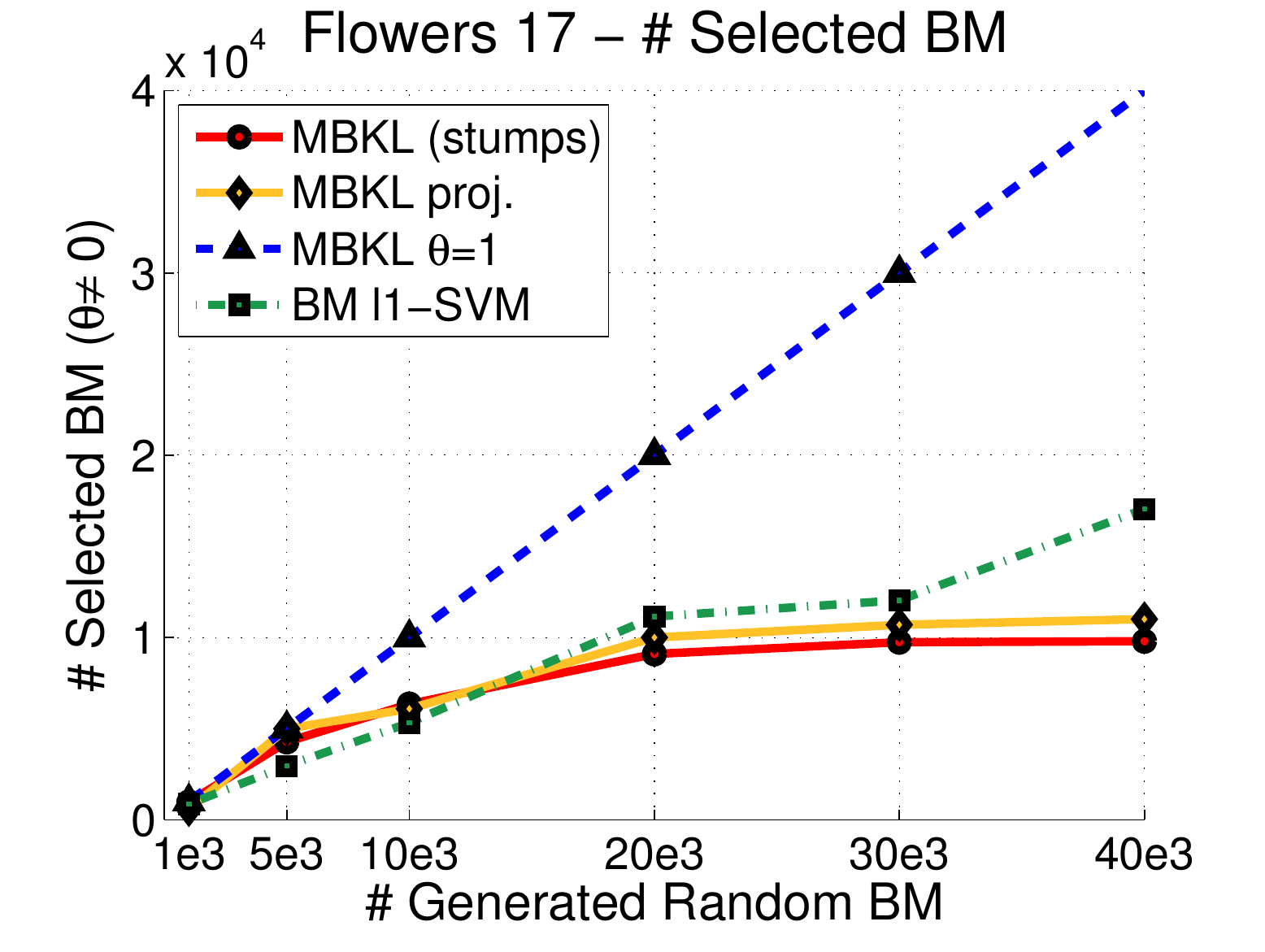}\\
 (c) & (d)\\
 \end{tabular} 
\caption{\emph{Analysis of MBKL in Flowers17.} (a) Accuracy, (b) training time,
(c) testing time, and (d) amount of selected BMs, when varying the amount of initially randomly generated BMs. 
}
\label{figtime}
\end{figure*}

\subsection{Analysis of  MBKL }
We investigate the impact of the different MBKL parameters.
Results  are given for Flowers17. We conducted the same analysis over the rest of the tested datasets (except ImageNet for computational reasons), and we 
could extract similar conclusions for all of them.
 To conduct our analysis we use the following baselines:
\vspace*{-0.1cm}
\paragraph{$-$ MBKL with binarized $\chi^2$ projections (MBKL proj.):} To test
 other BMs in MBKL, we replace the decision stumps by random projections.
We generate the random projections using the explicit feature map for 
the approximate $\chi^2$ by~\cite{Vedaldi11}. To obtain the BMs, we threshold them using decision stumps. 
We use the $\chi^2$ random projections because they achieve high performance on Flowers17, and the computational cost is of the 
same order as any other random projection. 
\vspace*{-0.1cm}
\paragraph{$-$ MBKL with $\btheta=\1$}: In order to analyze the impact of $\btheta$, we do not learn it
 and directly set it to $\1$. This is equivalent to using BMs as the input features to a linear SVM.
\vspace*{-0.1cm}
\paragraph{$-$ BMs as input features for $\ell_1$-SVM}: We use the BMs as the input features to SVM with a $\ell_1$ regularizer, which allows for 
discarding more BMs. Note that in contrast to MBKL, the selection of BMs is different for each class.

\vspace*{-0.1cm}
\paragraph{\bf Impact of the BMs.}
In Fig.~\ref{figtime}, for MBKL and the different baselines, we report
  the accuracy, training time, testing time, and the final amount of
BMs with $\theta_k\neq0$, when  varying the amount of initial BMs. 
Comparing random projections and decision stumps, we observe  that
both obtain similar performance. 
Also, note that each random projection has a computational complexity in the order of the feature length, $\bigO(n)$,
and decision stumps of $\bigO(1)$. This is noticeable at test time, and not in training,
since the learning algorithm is much more expensive than computing the BMs.

We can observe that when increasing the number of BMs, the accuracy saturates and does not degrade.
MBKL does not suffer from over-fitting when including a large amount of BMs. 
We believe that this is
 because the BMs are generated without labeled data, and then are used in a kernel SVM that is properly regularized.

\vspace*{-0.1cm}
\paragraph{\bf Impact of $\btheta$.}
In Fig.~\ref{figtime} we also show the results of fixing $\btheta=\1$. This yields a performance close to that of MBKL, 
because the SVM parameters compensate for the lack of learning $\btheta$. 
Note that fixing $\btheta=\1$ lowers the training time since the kernel need not be learned. 
Yet, learning the kernel 
is justified because it allows to discard BMs for all classes together (recall that the kernel
does not vary depending on the image class), which yields a faster testing time. 
The number of BMs diminishes after learning the kernel.
We can see that this is also the case for the BMs in an $\ell_1$ framework, which is efficient to evaluate, but degrades  the performance.

Interestingly, observe that the original feature length is 
$31,300$, and when using an initial amount of $5,000$ BMs, 
the performance is already very competitive. Also,  
note that the number of BMs with $\theta_k\neq 0$ saturates at around $10,000$ (Fig.~\ref{figtime}d).
This is because there are redundant feature descriptors, or non-informative feature pooling regions,
and MBKL learns that they are not  relevant for the classification. 
We only observed this drastic reduction of the feature length on Flowers17, where we use multiple descriptors.  

\begin{table}[t!]
\centering
\begin{tabular}{|@{\hspace{2mm}}l@{\hspace{3mm}}|@{\hspace{3mm}}c@{\hspace{3mm}}c@{\hspace{3mm}}c@{\hspace{3mm}}c@{\hspace{3mm}}c@{\hspace{3mm}}c|}

\hline
\emph{Datasets}	 & \bf{Daimler }  & \bf{Flowers17}&\bf{OSR}& \bf{a-VOC08}&\bf{ImgNet } & \bf{ImgNet }\\
\emph{Descr.}& HoG			& BoW 			& HoG+Attr				& HoG+Attr & 		BoW &  S.Quant.\\
\emph{Descr. Length} & $558$ &  $31e3$  &$518$ &$9,752$  &$21e3$ & $21e{3}$\\
\hline
\hline
\emph{\# Initial BMs} &$1e{4}$  & $3e{4}$&$3e4$ &$1e5$ & $ 2e{5}$& $21e{3}$\\
\emph{\# $\theta_k\neq0$} &$3,832$  &$9,740$ & $2,906$& $6e4$ & $9e4$ &$18,341$ \\
\emph{Step $1$: Neg/Pos} & $1$  & $10$&$7$ &$2$ & $2$ &$2$\\
\emph{Step $1$: Samples} &$19,600$  & $11,220$&$1,920$ &$19,020$ & $5e4$ & $5e4$\\
\emph{Step $1$ Time}& $20$s & $56$s & $5$s& $480$s& $  5e3$s&$ 2e3$s   \\
\emph{Total Time}& $24$s & $60$s & $5.5$s & $690$s& $2e5$s&$ 5e4$s   \\
\hline
 \end{tabular}
   \caption{ \emph{Learning Parameters of MBKL.}  We  report the amount of
BMs randomly generated (\# Initial BMs), the number of BMs selected (\# $\theta_k\neq0$), the proportion of Positive vs Negative training samples  (Neg/Pos),
 the amount of samples (Samples), and the training times for one split. }
   \label{taccTrainc}
\end{table}

\vspace*{-0.1cm}
\paragraph{\bf Computational Cost of Kernel Learning.} 
Table~\ref{taccTrainc} shows the impact of the parameters on the computational cost of the learning algorithm. 
We also report for all datasets the MBKL parameters that we use in the rest of the experiments. 
Recall that the parameters of Step~$1$ are the initial number of BMs and the number of training samples we use to 
 learn  the \emph{two-class} SVM. 
 We set these parameters to strike a good balance between accuracy and efficiency.
The proportion of negative vs. positive training samples is set to $2$, except in cases where this yields insufficient training data. 
We observe that the initial number of BMs  is usually $10$ to $100$ times the length of the image descriptor.
As a consequence,  learning $\btheta$ may become a computational bottleneck. 
Fortunately, the accuracy of Step~$1$ flattens out after a small number of selected training samples.

After learning the kernel,
the number of BMs with $\theta_k\neq 0$ is about $10$ times the length of the image descriptor. 
Thus, learning the final one-vs-all SVM (Step~2) usually is computationally cheaper then learning $\btheta$, though the cost increases with 
the number of classes.

\subsection{Comparison to state-of-the-art}
We compare MBKL with other learning methods based on binary decisions, and also with the state-of-the-art (s-o-a) efficient SVM methods. 
In all cases we use the code provided by the authors.
\vspace*{-0.1cm}
\paragraph{\bf Methods based on binary decision:} We compare to Random Forest (RF)~\cite{Breiman01} 
using 100 trees of depth 50, except for UCI  where we  use 50 trees of depth 10.  
We also compare to the AdaBoost implementation of~\cite{Vezhnevets05},
 using 500 iterations. These parameters  were the best found.

\vspace*{-0.1cm}
\paragraph{\bf Predefined kernels:} We use
 { $\chi^2$, RB-$\chi^2$ kernels}, and 
{Intersection kernel (IK)~\cite{Maji12}}. 
For RB-$\chi^2$ we set the hyper-parameter of the kernel to the mean of the data.

\vspace*{-0.1cm}
\paragraph{\bf Fast kernel approximations:} 
We use some of the state-of-the-art methods:
\begin{itemize}[leftmargin=0cm,itemindent=-.5cm,labelwidth=\itemindent,labelsep=.4cm,align=left]
\item[] \noindent \emph{- Approx. $\chi^2$ by Vedaldi and Zisserman~\cite{Vedaldi11}:} we use 
an expansion of $3$ times the feature length, which is reported in~\cite{Vedaldi11} to work best.
We also use an expansion of $9$  times, which gives a feature 
length similar to MBKL (we indicate this with: x$3$).

 \item[] \noindent \emph{- Approx. Intersection Kernel (IK) by Maji~\etal~\cite{Maji12}:} following the suggestion by the authors,  we use  $100$ bins for the quantization.
We did not observe any significative change in the  accuracy when further increasing the number of  bins for the quantization.

\item[] \noindent \emph{- Power mean SVM (PmSVM) by Wu~\cite{Wu12PmSVM}:} We use the $\chi^2$ approximation and the intersection approximation of~\cite{Wu12PmSVM}, with the default parameters. The features are scaled following the author's recommendation.
\end{itemize}

\begin{table}[t!]

\centering
\begin{tabular}{|@{\hspace{2mm}}l@{\hspace{2.5mm}}|@{\hspace{2.5mm}}c@{\hspace{2.5mm}}c@{\hspace{2.5mm}}c@{\hspace{2.5mm}}c@{\hspace{2.5mm}}c
@{\hspace{2.5mm}}c@{\hspace{2.5mm}}
c@{\hspace{2.5mm}}c@{\hspace{2.5mm}}|}

\hline
\emph{Dataset} & \bf{Daimler } & \bf{Liver} & \bf{Sonar}& 
\bf{Flower}&\bf{OSR}&  \bf{aVOC08}&
\bf{ImNet }&\bf{ImNet}\\

\emph{Descriptor}& \emph{HoG  } &\emph{Attr. } &\emph{Attr. }& 
\emph{BoW  }& \emph{HoG+At  }&  
\emph{HoG+At}&\emph{BoW }&\emph{SQ. }\\

\emph{Length}& \emph{ $558$}  &\emph{ $6$} &\emph{ $60$ }& 
\emph{$31e3$ }&\emph{ $518$}&  
\emph{$9752$ }&\emph{ $21e{3}$}&\emph{ $21e{3}$}\\
\hline
\hline
{MBKL} 						    			
 	& $96.3$ &$\mathbf{75.0}$&$\mathbf{86.3}$  	& $88.5$  	&$\mathbf{77.1}$&$62.1$&$22.3$&$26.1$\\
\hline
\hline
{Linear SVM}					    			
&  $94.1$ &$67.5$&$77.1$  & $64.6$  &$73.4$&$57.9$&$17.6$&$\mathbf{26.3}$\\
\hline
\hline
{R. Forest}					    	
 & $93.9$ &$73.0$&$79.5$  & $77.2$  &$73.6$&$46.6$&$-$&$-$\\
{AdaBoost}					    			
 & $93.7$ &$72.2$&$83.9$  & $61.1$  &$57.7$&$35.6$&$-$&$-$\\

\hline
\hline
{$\chi^2$}					
 & $96.2$ &$68.1$&$82.4$  & $87.4$  &$76.6$&$61.4$&$-$&$-$\\
{RB-$\chi^2$ }				
 & $\mathbf{96.6}$ &$70.7$&$82.4$& $85.9$ &$76.0$&$\mathbf{64.0}$&$-$&$-$\\
{Appr $\chi^2$ x$3$}
 & $96.2$ &$72.7$&$82.4$   & $87.2$  &$76.0$&$62.3$  & $-$  & $-$\\
{Appr $\chi^2$}
 & $96.1$ &$72.5$&$81.9$ & $87.2$  &$75.9$&$62.0$&$22.0$&$23.5$\\
{PmSVM $\chi^2$}
 & $93.2$ &$50.0$&$73.8$& $\mathbf{90.8}$ &$71.9$&$63.5$&$\mathbf{22.6}$&$23.7$\\
\hline
\hline
{IK}
&  $95.9$ &$73.3$&$84.9$& $86.6$  &$72.2$&$53.3$&$-$&$-$\\
{Appr IK}
& $95.6$ &$59.1$&$81.0$& $86.8$ &$\mathbf{77.1}$&$62.5$&$-$&$-$\\
{PmSVM IK}
 & $91.2$ & $50.0$ &$65.5$ & $90.6$ &$71.5$&$63.5$&$\mathbf{22.6}$&$23.9$\\

\hline
 \end{tabular}
   \caption{ \emph{Evaluation of the performance on all datasets.}  We  report the accuracy using the standard
evaluation setup for each dataset. }
   \label{tacc}
\end{table}

We also analyzed the use of BMs for kernel approximation by Raginsky and Lazebnik~\cite{Raginsky09} (using the available code). 
This approach combines BMs and the random projections by Rahimi and Recht~\cite{Rahimi07}. In contrast to MBKL, 
that method learns the kernel distance to preserve the locality of the original descriptor space. 
We use the resulting kernel in a linear SVM, and 
we found that it performs poorly (we do not report it in the tables). 
Note that this method was 
designed to preserve the locality, which is a useful criterion for image retrieval but may be 
less so for image classification. Moreover, it is based on the random projections of~\cite{Rahimi07} 
that approximate the RBF kernel, which might not be adequate for the image descriptors we use.

We do not report the accuracy performance of the MKL method for large-scale data by Bazavan \etal~\cite{Bazavan12} (for which the code is not available). 
 This is because~\cite{Bazavan12} uses the approximations of the 
predefined kernels that we already report, and the accuracy very probably is 
comparable to those approximations with the parameters set by cross-validation.

\vspace*{-0.2cm}
\paragraph{\bf Performance accuracy.} 
The results are reported in Table~\ref{tacc}. We can observe that MBKL is the only method that 
for each benchmark achieves an accuracy similar to the best performing method for that benchmark.
Note that the kernels and their approximations do not perform equally well for all descriptor 
types. Their performance may degrade when the descriptors are attribute-based features, and also, 
when descriptors are already linearly separable, such as s-o-a descriptors in large scale datasets.
In most cases, the approximations to a predefined kernel perform similarly to the actual kernel. 
We observed that the performance of PmSVM is lower when the feature length is small. We can 
conclude MBKL outperforms the other methods based on binary decisions, including Random Forest 
and Boosting. 

For all tested datasets the accuracy of MBKL is comparable to the s-o-a, which is normally only
achieved by using different methods for different datasets. We even outperform the s-o-a for UCI~\cite{Gai2010}, as 
well as for
the Daimler~\cite{Maji12} benchmark since we found better parameters for the HoG features. For ImageNet, MBKL outperforms~\cite{Boix12SQ} using the same
descriptors, achieving a good compromise between accuracy and efficiency (computing the 
descriptors for the whole dataset in less than 24h using 4 CPUs).

\begin{figure}[t!]
\centering
\begin{tabular}{c@{\hspace{0.2cm}}c@{\hspace{0.2mm}}c}
\multicolumn{3}{c}{ \includegraphics[scale=0.29]{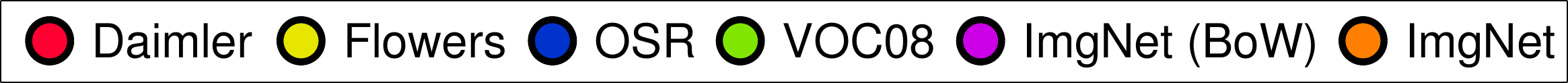}}\\

 \includegraphics[scale=0.245]{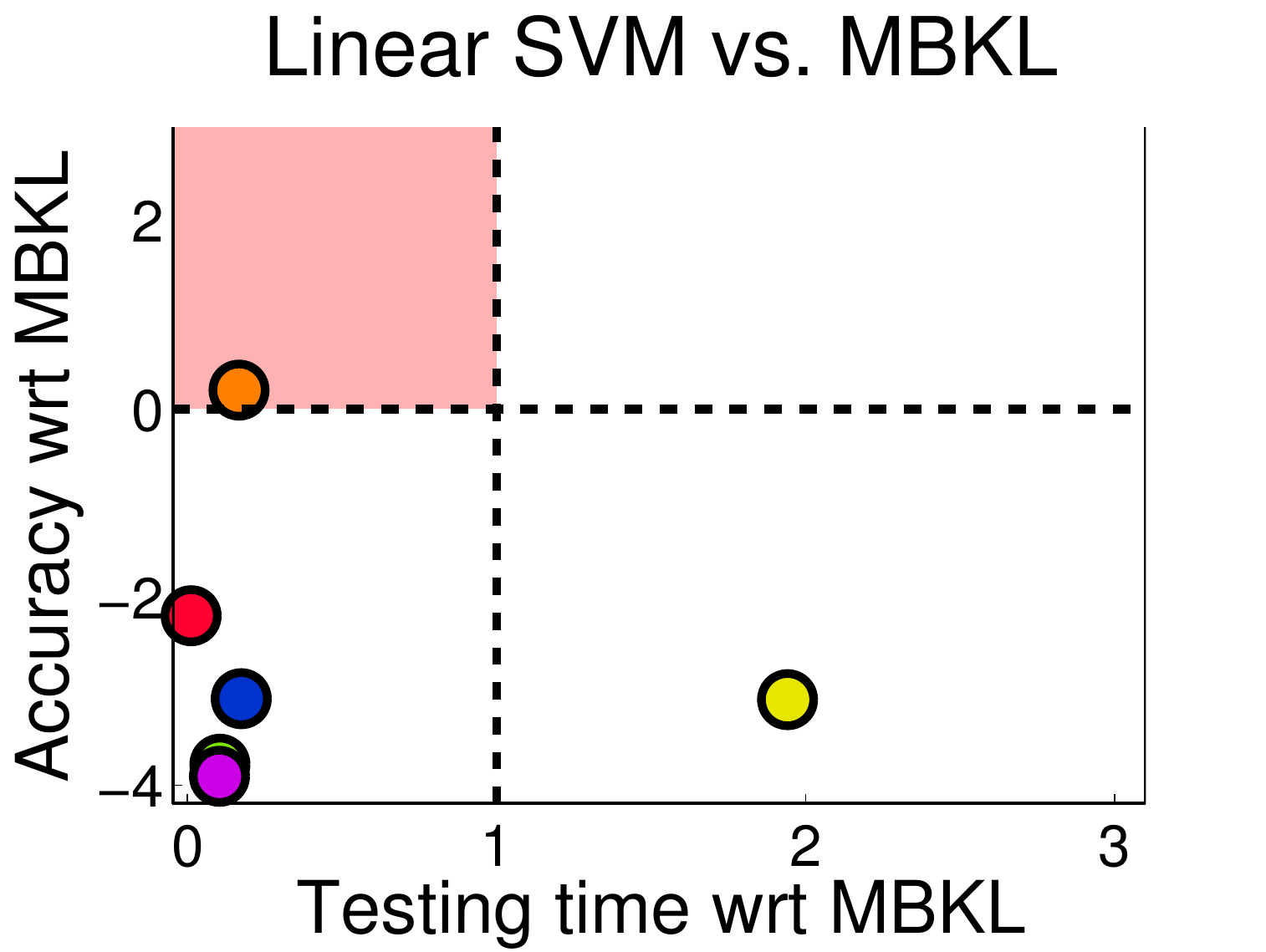} &  \includegraphics[scale=0.25]{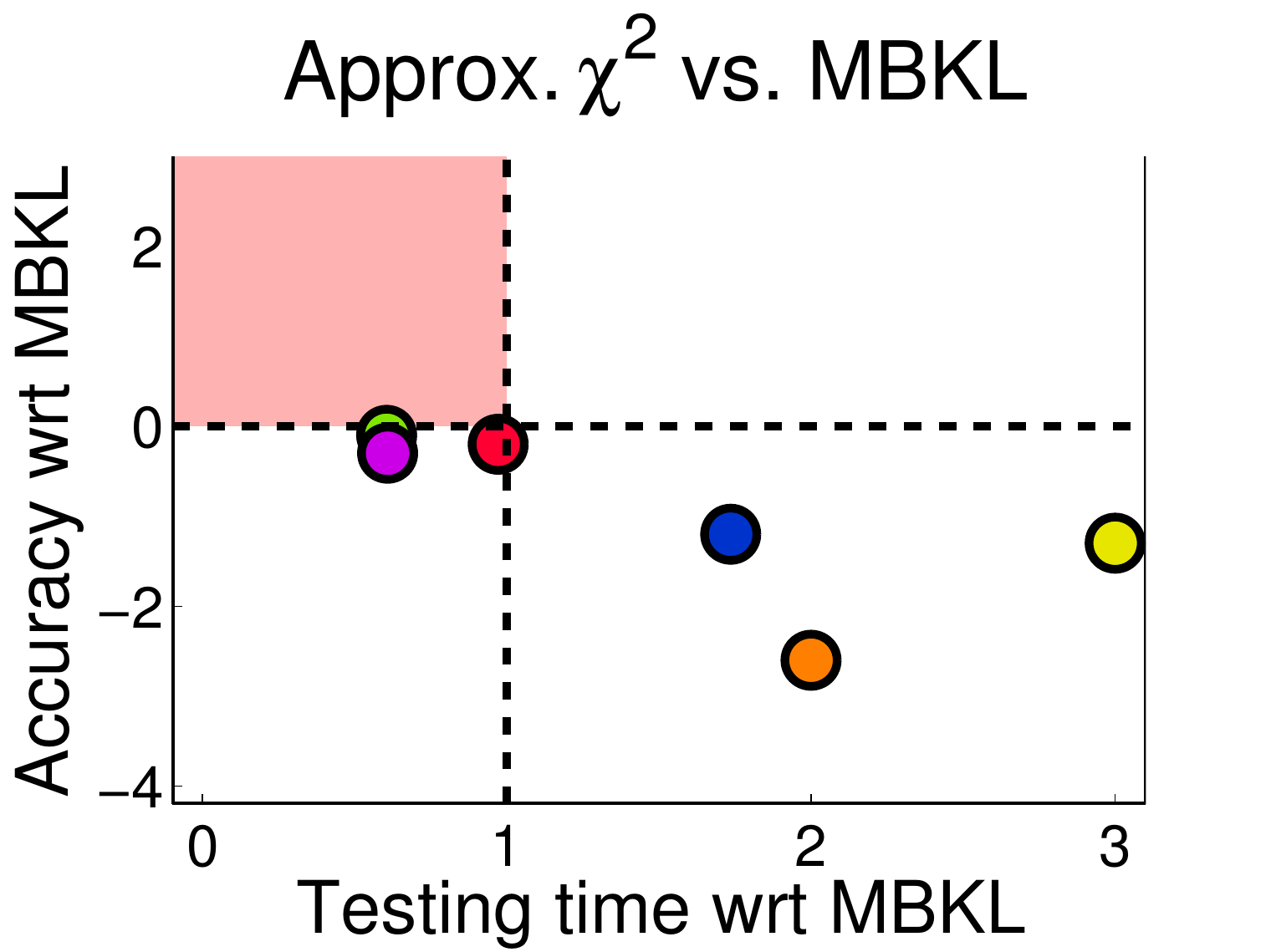}&  \includegraphics[scale=0.25]{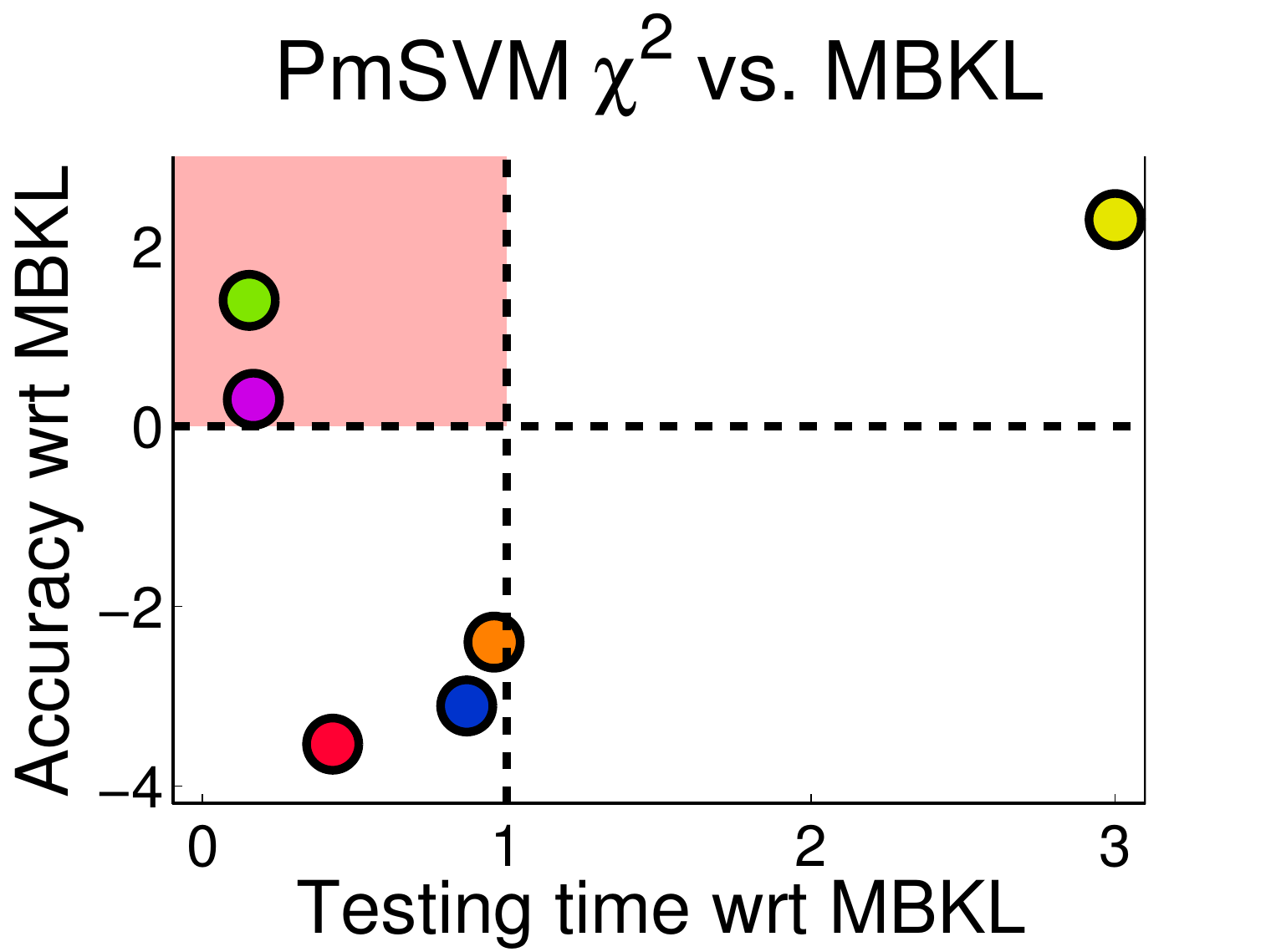}\\
(a) & (b) & (c) \\
\end{tabular}
\caption{\emph{Testing Time and Accuracy in all Datasets.} The 
computational cost is normalized with respect to MBKL. A score of $2$ means that is two times slower than MBKL.
The accuracy with respect to MBKL is the difference between the accuracy of the competing method and MBKL. 
A score of $2$ means the other method performs $2$\% better
than MBKL. Points in the red area indicate that the other method perform better than MBKL for both speed and accuracy.}
\label{figba}
\end{figure}

\vspace*{-0.2cm}
\paragraph{\bf Test Time.} Fig.~\ref{figba} compares the testing time of MBKL to that of the 
efficient SVM methods. We report the testing times relative to the test time of MBKL, as well as the
accuracy relative to MBKL. 
MBKL achieves very competitive levels of efficiency. 
MBKL's computation speed depends on the number of BMs with $\theta_k\neq 0$. For Flowers17, 
MBKL is faster than linear SVM because there are fewer final BMs than original feature 
components. Note also that if the final length of the feature map is the same for MBKL and 
approximate $\chi^2$, MBKL can be faster because decision stumps are faster to calculate than 
the projections to approximate $\chi^2$. PmSVM achieves better accuracy and speed than MBKL in two cases, but note
that for the rest of the cases this is opposite. Note that in UCI datasets, which are not Fig.~\ref{figba} but in Table~\ref{tacc},  
PmSVM performs poorly compared to MBKL, since the descriptors are  attribute-based.

\vspace*{-0.2cm}
\paragraph{\bf Training Time.} 
When learning MBKL, any of the optimizations can be done with an {off-the-shelf} linear SVM. 
Thus, the computational complexity depends mainly on the optimizer. We use liblinear, but we 
could use any other more efficient optimizer. Comparing the different SVM optimizers is out 
of the scope of the paper. We do not compare the training time of MBKL to that of the methods 
that use predefined kernels, because these methods do not have the computational overhead of 
learning the kernel.

\vspace{-0.1cm}
\section{Conclusions}
\label{sec:conclusion}
This paper introduced  a new kernel that is learned by combining a large amount of simple, randomized BMs.
We derived the form of the non-linear mapping of the kernel that allows similar levels of efficiency to be reached as the fast kernel SVM approximations. 
Experiments show that our learned kernel can adapt to most common image descriptors, 
achieving a performance comparable to that of kernels specifically selected for each image descriptor.
We expect that the generalization capabilities of our kernel can
be exploited to design new, unexplored, image descriptors.

\bibliographystyle{splncs}
\bibliography{egbib}

\end{document}